\documentclass[runningheads]{llncs}
\usepackage{graphicx}
\usepackage{amsmath}
\usepackage{amssymb}
\usepackage[linesnumbered,ruled,vlined]{algorithm2e}
\usepackage{subfigure}
\usepackage{algpseudocode}

\begin{document}
\title{An LGMD Based Competitive Collision Avoidance Strategy for UAV\thanks{The first and second author contributed equally. This research is funded by the EU HORIZON 2020 project, STEP2DYNA (grant agreement No. 691154) and ULTRACEPT (grant agreement No. 778062)}}
%
%
\author{Jiannan Zhao\inst{1}\orcidID{0000-0003-0052-3365} \and
Xingzao Ma\inst{2}\orcidID{0000-0001-7299-6677} \and
Qinbing Fu\inst{1}\orcidID{0000-0002-5726-6956} \and
ChengHu \inst{3}\orcidID{0000-0002-1177-2167} \and
Shigang Yue\inst{1}\orcidID{0000-0002-1899-6307}
}

\authorrunning{J. ZHAO et al.}
%
\institute{University of Lincoln, Brayfordpool LN6 7TS, UK \and
College of Electro-mechanical and Engineering, Lingnan Normal University, Zhanjiang 524048, China\and
Machine Life and Intelligence Research Centre, Guangzhou University, China
\email{maxz@lingnan.edu.cn}\\
\email{\{jzhao,qfu,chu,syue\}@lincoln.ac.uk}}
\maketitle              
\begin{abstract}
Building a reliable and efficient collision avoidance system
for unmanned aerial vehicles (UAVs) is still a challenging problem. This
research takes inspiration from locusts, which can fly in dense swarms for
hundreds of miles without collision. In the locust's brain, a visual path-
way of LGMD-DCMD (lobula giant movement detector and descending
contra-lateral motion detector) has been identified as collision perception system guiding fast collision avoidance for locusts, which is ideal
for designing artificial vision systems. However, there is very few works
investigating its potential in real-world UAV applications. In this pa-
per, we present an LGMD based competitive collision avoidance method
for UAV indoor navigation. Compared to previous works, we divided the
UAV's field of view into four subfields each handled by an LGMD neuron.
Therefore, four individual competitive LGMDs (C-LGMD) compete for guiding the directional
collision avoidance of UAV. With more degrees of freedom compared to
ground robots and vehicles, the UAV can escape from collision along four
cardinal directions (e.g. the object approaching from the left-side triggers
a rightward shifting of the UAV). Our proposed method has been validated by both simulations and real-time quadcopter arena experiments.

\keywords{UAV Collision Avoidance, LGMD, Bio-inspired Neural Network}
\end{abstract}
\section{Introduction}
UAV is one of the most attractive but vulnerable robot platform, which has the potential to be applied in tons of scenarios, such as geography survey, agriculture fertilization, exploration in dangerous or disaster regions, products delivery, shooting photos. Safety of the UAV is always a vital property in a UAV application. Thus, researchers always seeking for better Sense and Avoidance (SAA) technics for UAVs. Classic UAVs use GPS or optic flow\cite{honegger2013open,sabo2016bio} to navigate, and onboard distance sensor like ultra sonic, infrared, laser, or a cooperative system to avoid obstacles as reviewed by\cite{yu2015sense}. However, these distance sensors are largely dependent on obstacles' materials, texture and backgrounds' complexity, thus, they can only work in simple and structured environment\cite{chee2013control}. Lidar and Vision based methods is more diverse and applicable. One popular vision based method is to detect and locate obstacles in a reconstructed map, mark out the frontiers of the obstacles as banded fields in the map, and then use specified pathfinding algorithm (e.g. heuristic algorithm) to generate safe trajectories to avoid collision \cite{richter2016polynomial,bachrach2009autonomous,achtelik2009stereo}.
This is also named as Simultaneous Localization and Mapping(SLAM), but its high demand of computational burden prevent it from small or micro UAVs.

On the other hand, bio-inspired vision based collision detecting methods are standing out for their efficiency. For example, Optic Flow (OF) is a widely used vision based motion detecting method inspired by biological mechanism in flies and bees\cite{serres2017optic}. It is also introduced in collision avoidance technology, e.g. Zufferey\cite{zufferey2005toward} applied 1D OF sensor onto a 30g light weight fixed wing UAV and achieved automatic obstacle avoidance in indoor(GPS denied) structured environment. And later in 2009\cite{beyeler2009vision} their group achieved autonomous avoidance towards trees with 7 OF sensor on a fixed wing platform. Griffiths\cite{griffiths2007obstacle} used optical mouse (key-point matching) converted OF sensor to fly through Canyon, besides the OF sensor, it also integrated a laser ranger for directly approaching obstacles. Serres\cite{serres2008vision} used a pair of EMD based OF sensor to avoid lateral obstacles for a hovercraft. Sabo\cite{sabo2016bio} applied OF onto quadcopters and repeated some benchmark experiments to analysis the behaviours for honeybee-like flying robot, however, the algorithm was still computed off board. Stevens\cite{stevens2018vision} achieved collision avoidance in cluttered 3D environments.

Lobula Giant Movement Detector (LGMD) is another bio-inspired neural network inspired by Locusts vision system, and especially, superior in detecting approaching obstacles and avoiding imminent collisions. Compared to Optic Flow, LGMD is more specialised for detecting directly approaching obstacles and eliminate redundant image difference caused by shifting things and backgrounds. The LGMD neuron and its presynaptic neural network has been modeled\cite{rind1996neural} and promoted by many researchers\cite{fu2017collision,fu2018shaping,yue2006collision}. As a collision detecting model, LGMD has been introduced to mobile robots\cite{hu2016bio}\cite{Sergi2010Strider}, embedded systems\cite{fu2016bio,hu2014development}, hexapod walking robot\cite{vcivzek2017neural},blimp\cite{Sergi2007fly} and cars\cite{yue2006bio,yue2007synthetic}.

Basic LGMD model provide the threat level of collision in the whole field of view (FoV), but it is not enough to make wise avoidance behaviour, hence, early research generated randomly turn direction in mobile robots\cite{hu2016bio}. Shigang\cite{Yue2010Reactive} divided the field of view into two bilateral halves, and discussed both winner-take-all and steering-wheel network in direction control system of the mobile robot. Compared to mobile robots, UAV has more degree of freedom, and is more vulnerable during flight. In the extremely limited literature of LGMD research on UAV platforms, Salt\cite{salt2017obstacle} implemented a neuromorphic LGMD model using recording from a UAV platform, and divided the FoV into half twice for direction information. But there is no real-time flight conducted. Our previous research has proved the applicability of LGMD on Quadcopter\cite{zhao2018bio} real-time flight and collision avoidance. Previously, the quadcopter can only avoid obstacles by randomly turn left or right in horizontal plane. To acquire the information about the coming direction of imminent obstacles, this research proposed a new image partition strategy, especially for LGMD application on UAVs, and a corresponding steering method for 3D avoidance behaviour. Both video simulation and real-time flight demonstrated the performance of this method.

\section{Model Description}
\subsection{LGMD Process}
\begin{figure}
  \centering

  \includegraphics[width=1.8in]{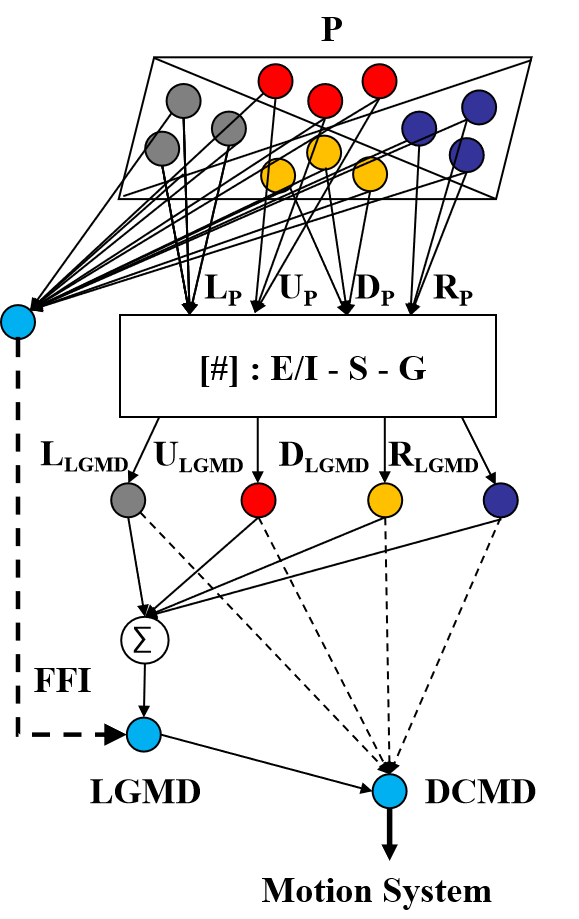}

  \caption{A schematic illustration of the proposed LGMD based competitive neuron network for collision detection. $[\#]$ denotes the inherited LGMD process as described in our previous research\cite{zhao2018bio}.}
  \label{fig:schematic}
\end{figure}

The LGMD process algorithm used in this paper is inherited from our previous research\cite{zhao2018bio}. The LGMD process is composed of five groups  of cells, which are P-cells (photoreceptor), I-cells (inhibitory), E-cells (excitatory), S-cells (summing) and G-cells (grouping), compared to previous model, we added four single competitive LGMD cells representing LGMD output of four sections: Left, Right, Up, and Down. The image is divided as shown in Fig.\ref{fig:schematic}.


The first layer of the neuron network is composed of P cells, which are arranged in a matrix, formed by luminance change between adjacent frames. The output of a P cell is given by:
\begin{equation}\label{qt:1}
  P_{f}(x,y)=L_f(x,y)-L_{f-1}(x,y)
\end{equation}
where $P_f(x,y)$ is the luminance change of pixel$(x,y)$ at frame $f$, $L_f(x,y)$ and $L_{f-1}(x,y)$ are the luminance at frame $f$ and the previous frame.

The output of the P cells forms the input of the next layer and is processed by two different types of cells, which are I (inhibitory) cells and E (excitatory) cells. The E cells pass the excitatory flow directly to S layer so that the E cells has the same value to its counterpart in P Layer; While the I cells pass the inhibitory flow convoluted by surrounded delayed excitations.

The I layer can be described in a convolution operation:
\begin{equation}\label{qt:2}
  [I]_f=[P]_f\otimes[w]_I
\end{equation}
where $ [w]_I $is the convolution mask representing the local inhibiting weight distribution from the centre cell of P layer to neighbouring cells in S layer, a neighbouring cell's local weight is reciprocal to its distance from the centre cell. To adapt fast image motion during UAV flight, $ [w]_I $is set differently to it in mobile robot\cite{hu2016bio}, the inhibition radius is expanded to 2 pixels:
\begin{equation}\label{qt:3}
  [w]_I=0.25\begin{bmatrix}
              \frac{1}{\sqrt{8}} & \frac{1}{\sqrt{5}} & \frac{1}{2} & \frac{1}{\sqrt{5}} & \frac{1}{\sqrt{8}} \\
              \frac{1}{\sqrt{5}} & \frac{1}{\sqrt{2}} & 1           & \frac{1}{\sqrt{2}} & \frac{1}{\sqrt{5}} \\
              \frac{1}{2}        & 1                  & 0           & 1                  & \frac{1}{2} \\
              \frac{1}{\sqrt{5}} & \frac{1}{\sqrt{2}} & 1           & \frac{1}{\sqrt{2}} & \frac{1}{\sqrt{5}} \\
              \frac{1}{\sqrt{8}} & \frac{1}{\sqrt{5}} & \frac{1}{2} & \frac{1}{\sqrt{5}} & \frac{1}{\sqrt{8}}
            \end{bmatrix}
\end{equation}

The next layer is the Sum layer, where the excitation and inhibition from the E and I layer is combined by linear subtraction, and after summation. Next, Group layer is involved to reduce the noise caused by sporadic image change or backgrounds. Detailed equation and parameters can be found in our previous work\cite{zhao2018bio}.

\begin{figure}
  \centering

  \includegraphics[width=2.2in]{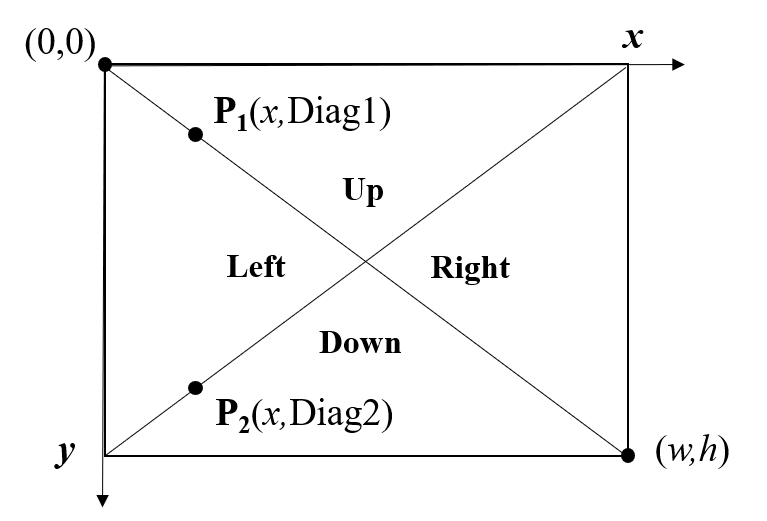}

  \caption{Image dividing method. The image scene is split through the diagonal. }
  \label{fig:Segmentation}
\end{figure}

When it comes to G layer, The unnormalized membrane potential of four C-LGMDs are Calculated respectively:
\begin{align}\label{qt:4567}
& U_\_{LGMD0}=\sum_{x}^{}\sum_{y=0}^{min[Diag1,Diag2]}|\widetilde{G}_f(x,y)| \\
& D_\_{LGMD0}=\sum_{x}^{}\sum_{y=0}^{max[Diag1,Diag2]}|\widetilde{G}_f(x,y)| \\
& L_\_{LGMD0}=\sum_{x}^{}\sum_{y>=Diag1}^{y<=Diag2}|\widetilde{G}_f(x,y)| \\
& R_\_{LGMD0}=\sum_{x}^{}\sum_{y=>Diag2}^{y<=Diag1}|\widetilde{G}_f(x,y)| \\
\end{align}
where $Diag1$, $Diag2$, denote the coordinates in y axis of the two diagonals, and $G_f(x,y)$ is the cells value of G layer, as illustrated in Fig.\ref{fig:Segmentation}. For more details about the process from $P_{f}(x,y)$ to $G_f(x,y)$ please looks in our previous work\cite{zhao2018bio}.

Previously, the membrane potential of the LGMD cell $K_{f0}$ is the summation of every pixel in G layer:
\begin{equation}\label{qt:8}
  K_{f0}=\sum_{x}^{}\sum_{y}^{}|\widetilde{G}_f(x,y)|
\end{equation}

Now it also equals to the summation of the four C-LGMD neurons:
\begin{equation}\label{qt:9}
  K_{f0}=U_{LGMD}+D_{LGMD}+L_{LGMD}+R_{LGMD}
\end{equation}

and then $K_f$ is adjusted in range $(0,255)$ by a sigmoid equation:
\begin{equation}\label{qt:10}
  \kappa_f=\frac{\text{tanh}(\sqrt{K_{f0}}-n_{cell}C_1 )}{n_{cell}C_2} \times 255
\end{equation}
where $C_1$ and $C_2$ are constants to shape the normalizing function, limiting the excitation $\kappa_f$ varies within [0, 255], $n_{cell}$ represents the total number of pixels in one frame of image. The membrane potential of the four C-LGMDs, is also limited in $(0,255)$ by calculating their proportion in $K_{f0}$, instead of modified with sigmoid function again:
\begin{align}\label{qt:11_14}
& U_\_{LGMD}=\frac{U_{LGMD0}}{K_{f0}}\times \kappa_f \\
& D_\_{LGMD}=\frac{D_{LGMD0}}{K_{f0}}\times \kappa_f \\
& L_\_{LGMD}=\frac{L_{LGMD0}}{K_{f0}}\times \kappa_f \\
& R_\_{LGMD}=\frac{R_{LGMD0}}{K_{f0}}\times \kappa_f \\
\end{align}

If $\kappa_f$ exceeds its threshold, then an LGMD spike is produced:
\begin{equation}\label{qt:15}
  S_f^{spike}=\begin{cases}
                1, & \mbox{if } \kappa_f \geqslant T_s \\
                0, & \mbox{otherwise}.
              \end{cases}
\end{equation}
An impending collision is confirmed if successive spikes last consecutively no less than $n_{sp}$ frames:
\begin{equation}\label{qt:16}
  C_f^{LGMD}=\begin{cases}
               1, & \mbox{if } \sum\limits_{f \!- \! n_{sp}}\limits^{f} S_f^{spike} \geqslant{n_{sp}} \\
               0, & \mbox{otherwise}.
             \end{cases}
\end{equation}
And then, based on the result of the competitive C-LGMDs, DCMD will switch to the corresponding escape command, and the command is sent through USART interface to the flight control system. The process from DCMD to PID based motor control system is shown in  pseudocode Algorithm 1.

\begin{algorithm}[ht]
\label{alg:B}
  \caption{Escape direction steering algorithm}
  \KwIn{four competitive C-LGMDs refers to stimulus in the FoV: $U_{LGMD}$, $D_{LGMD}$, $L_{LGMD}$, $R_{LGMD}$}
  \KwOut{expected quadcopter speed in x,y,z axis (PID motor control input): $exp\_sp[x], $ $exp\_sp[y], $  $exp\_sp[z]$}

 \While{$C_f^{LGMD} = 1$}
 {
  $minDirection \gets U_{LGMD}$
  \\
    \If{$minDirection \ge D_{LGMD}$}
  {
      $minDirection\gets D_{LGMD}$
  }
    \If{$minDirection \ge L_{LGMD}$}
  {
      $minDirection\gets L_{LGMD}$
  }
      \If{$minDirection \ge R_{LGMD}$}
  {
      $minDirection\gets R_{LGMD}$
  }

  \If{$minDirection=U_{LGMD}$}{
   set $exp\_sp[z] = speed_0$;
    }
  \textbf{else}
  \If{$minDirection = D_{LGMD}$}{
   set $exp\_sp[z] = -speed_0$;
   }
  \textbf{else}
  \If{$minDirection = L_{LGMD}$}{
   set $exp\_sp[y] = speed_0$;
   }
   \textbf{else}
  \If{$minDirection = R_{LGMD}$}{
   set $exp\_sp[y] = -speed_0$;
   }
 }
\end{algorithm}\label{Algrthm_Motion}

\section{System Overview}
In this section, the outline of the whole system is described. A system composed of Quadcopter, embedded LGMD detector, Ground Station, Remote and auxiliary sensors is depicted in Fig.\ref{fig:Modules} .
Luminance information is collected by the camera on the detector board, and then involved into the LGMD algorithm, the output command is passed through a USART port into the flight control to monitor avoiding tasks.
\subsection{Quadcopter Platform}
\begin{figure}
  \centering
  \includegraphics[width=3.5in]{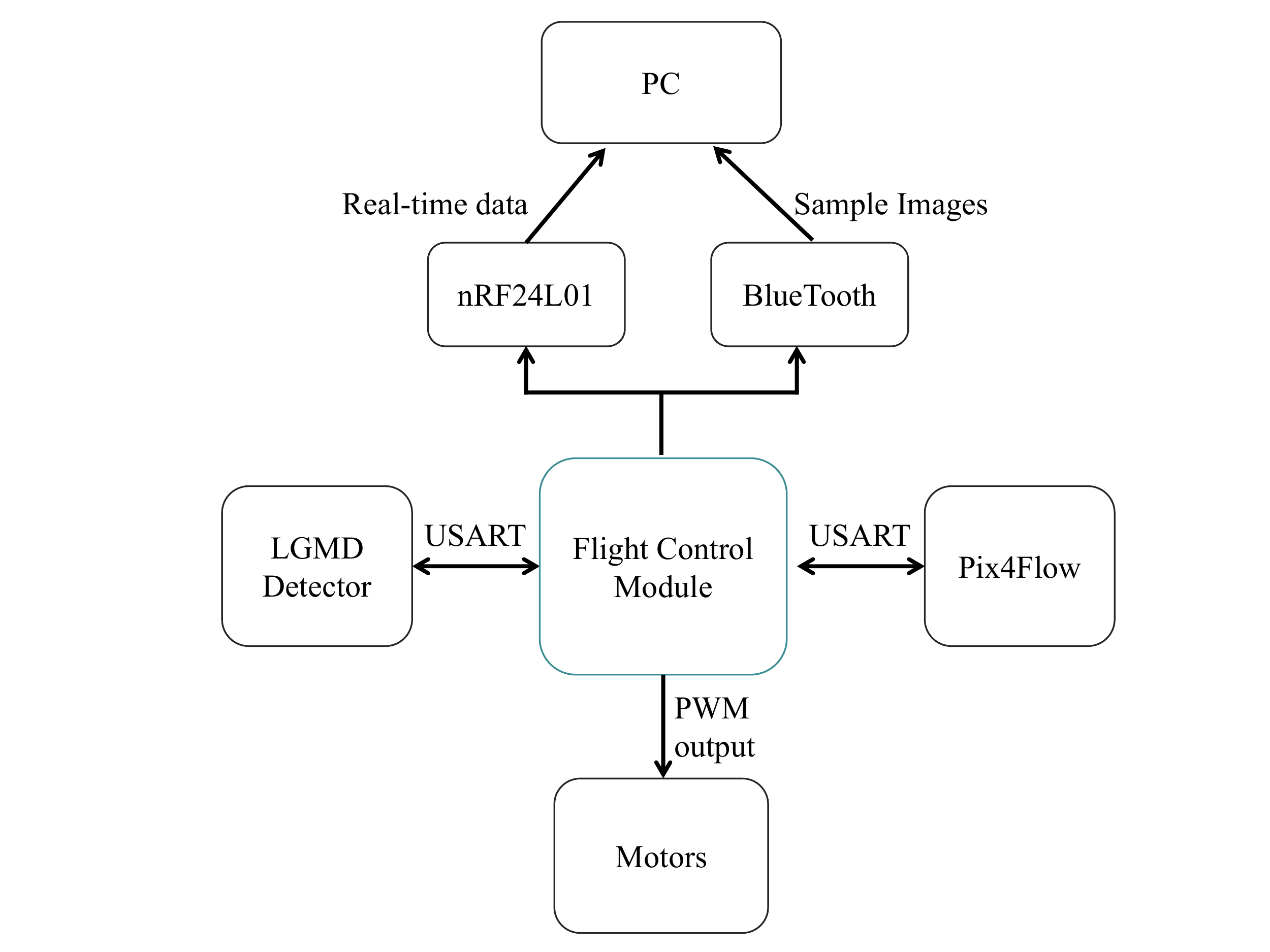}
  \caption{The structure of the quadcopter platform.}\label{fig:Modules}
\end{figure}

The UAV platform used in this research is a customized quadcopter with the skeleton size of 33cm between diagonally rotors. The flight control module we used is based on a STM32F407V and provides 5 USART interface for extra peripheral.
Multiple sensors are applied for data collection and enhance the stability of the quadcopter, including an IMU(Inertial Measurement Unit), an ultrasonic sensor, an optic flow sensor and the LGMD detector, as illustrated in Fig.\ref{fig:Modules}. The Pix4Flow optic flow module\cite{honegger2013open} is occupied as a position and velocity feedback in horizontal plane. The flight control module works as the central controller to combine the other parts together. It receives source data from the embedded IMU module(MPU6050), the Pix4flow optic flow sensor, and the LGMD detector, calculates out the PWM(Pulse-Width Modulation) values as the output to the four motors and it also sends back real time data for analysis through the nRF24L01 module.

\section{Experiments and Results}
To verify the performance of the proposed algorithm, both video simulation and arena real-time flight are conducted.
\subsection{Video Simulation}
\begin{figure}
\begin{minipage}{7cm}
\centering 
    \subfigure[]
    {
      \includegraphics[width=2.2in]{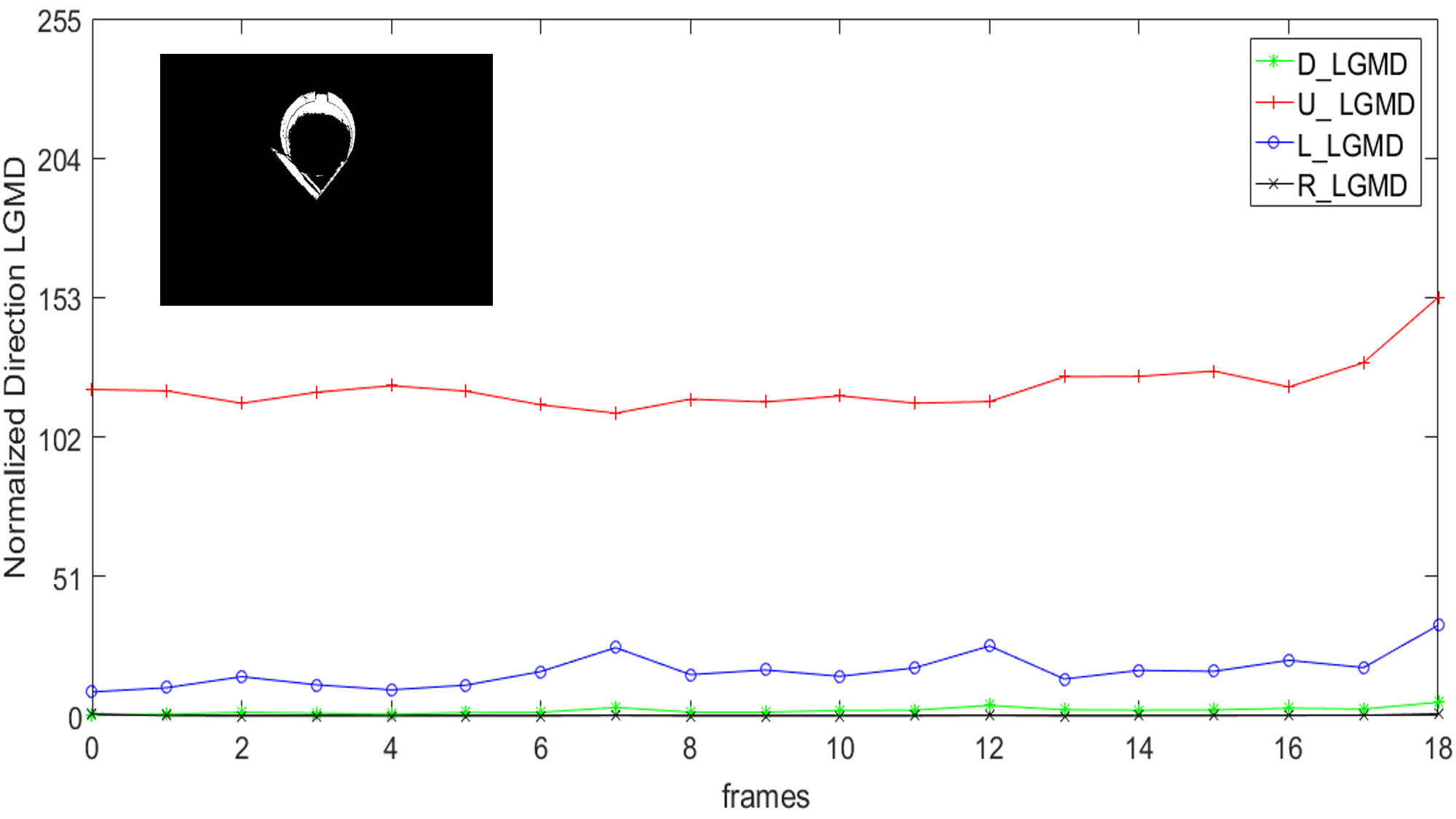}
    }
\end{minipage}
\begin{minipage}{7cm}
\centering 
    \subfigure[]
    {
      \includegraphics[width=2.2in]{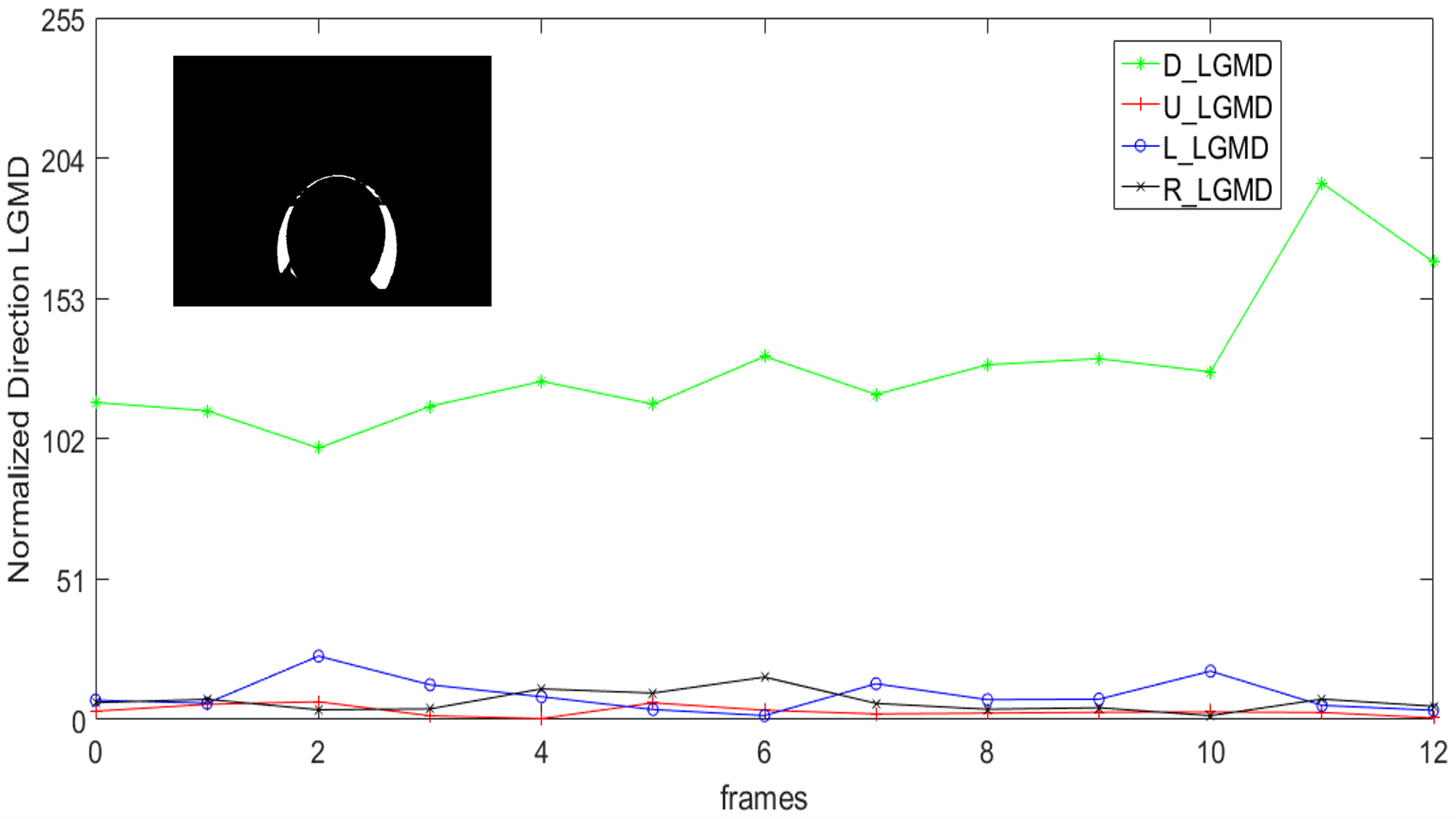}
    }
\end{minipage}
\begin{minipage}{7cm}
\centering 
    \subfigure[]
    {
      \includegraphics[width=2.2in]{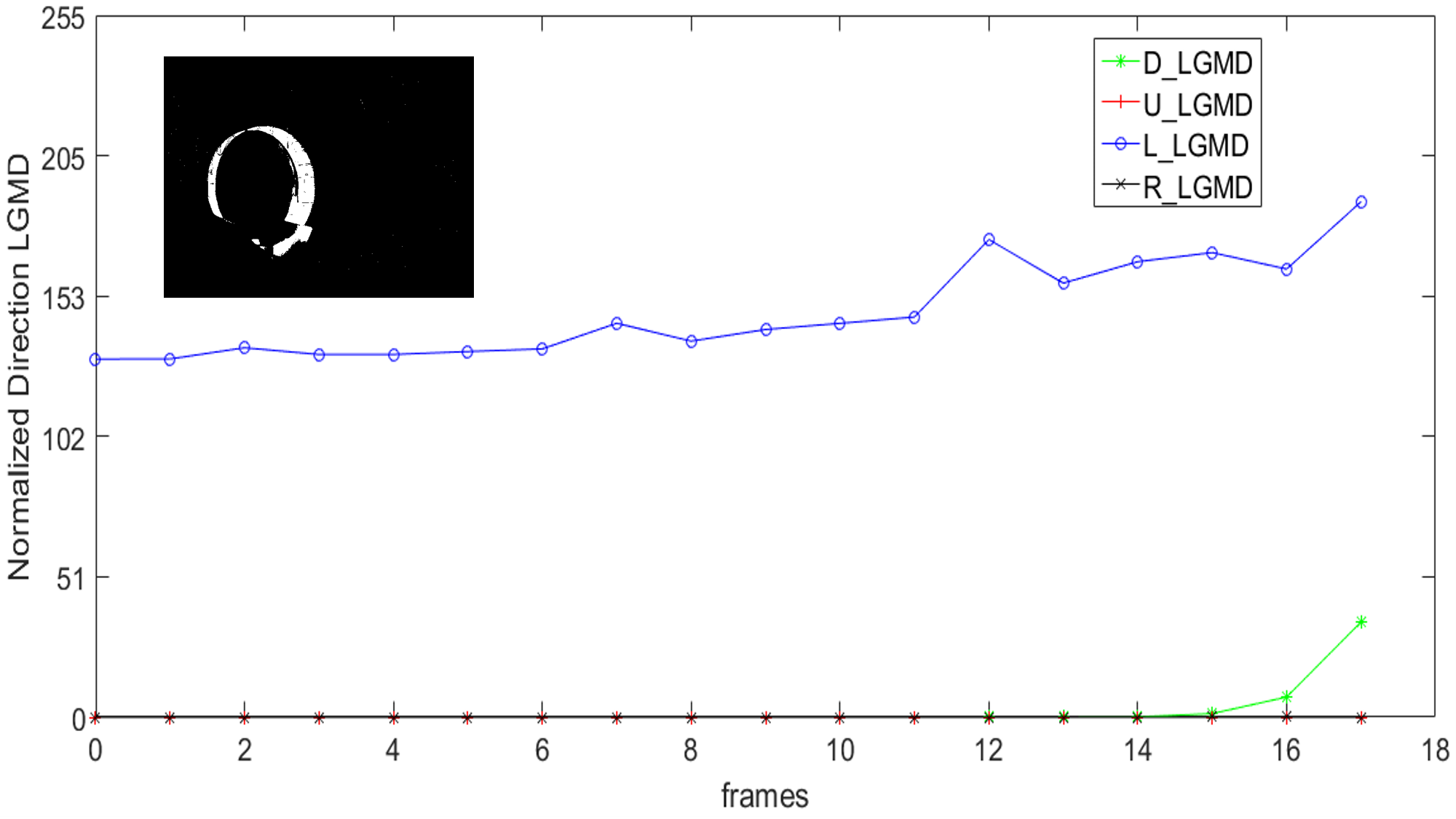}
    }
\end{minipage}
\begin{minipage}{7cm}
\centering 
    \subfigure[]
    {
      \includegraphics[width=2.2in]{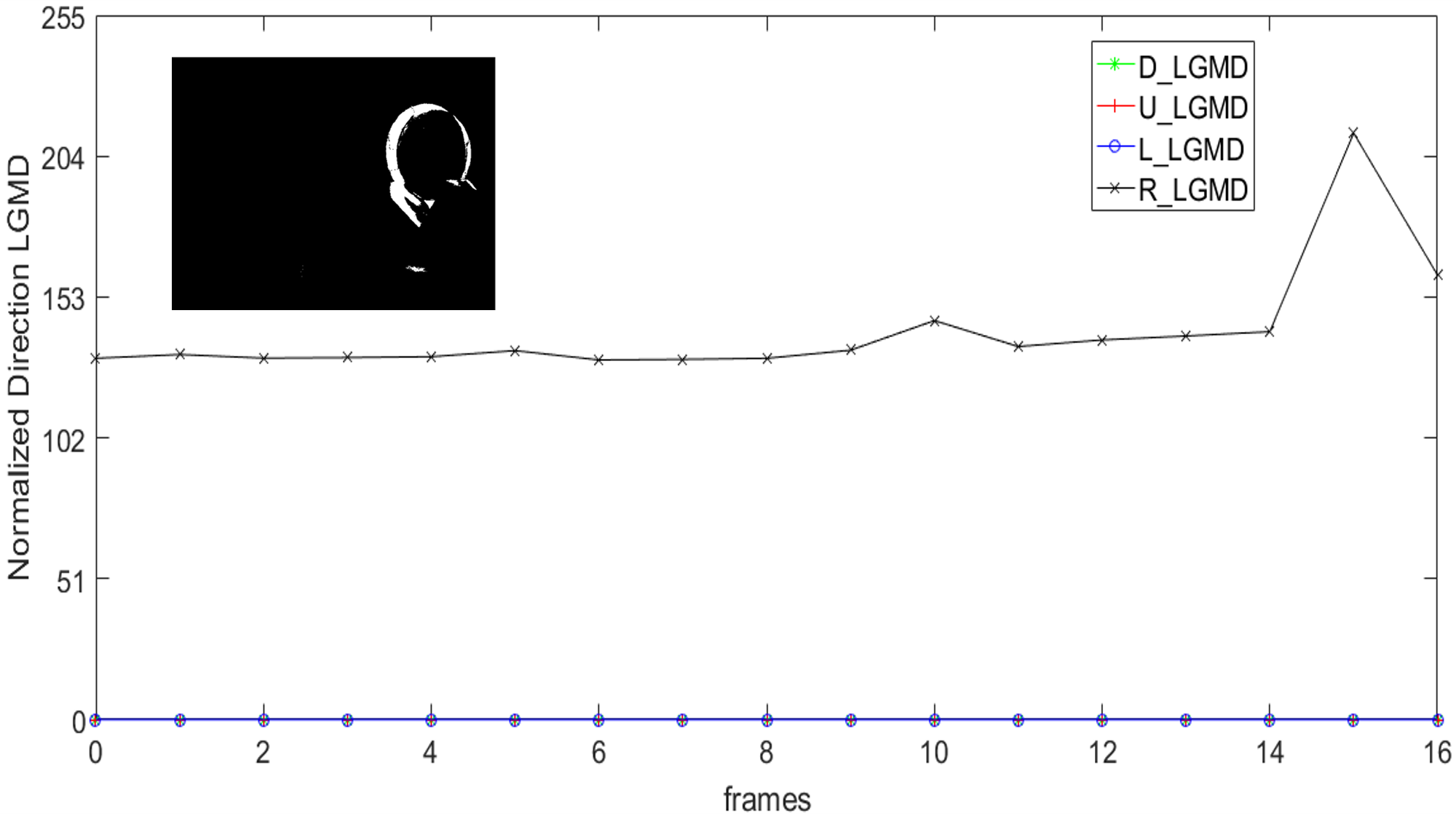}
    }
\end{minipage}
  \caption{Simulation results with snapshot.(a),(b),(c),(d) are membrane potential in C-LGMDs, toward upside, downside, left side, and right side stimuli respectively.}\label{fig:Simulation}
\end{figure}

The algorithm is firstly implemented on matlab and tested by a series recorded video, to verify whether the algorithm can distinguish stimulus from different directions. The results in Fig.\ref{fig:Simulation} indicate that the new network is able to respond differently towards coming objects from different directions.

\subsection{Hovering \& Features Analysis}
\begin{figure}[htbp]
\centering
\includegraphics[width=3.0in]{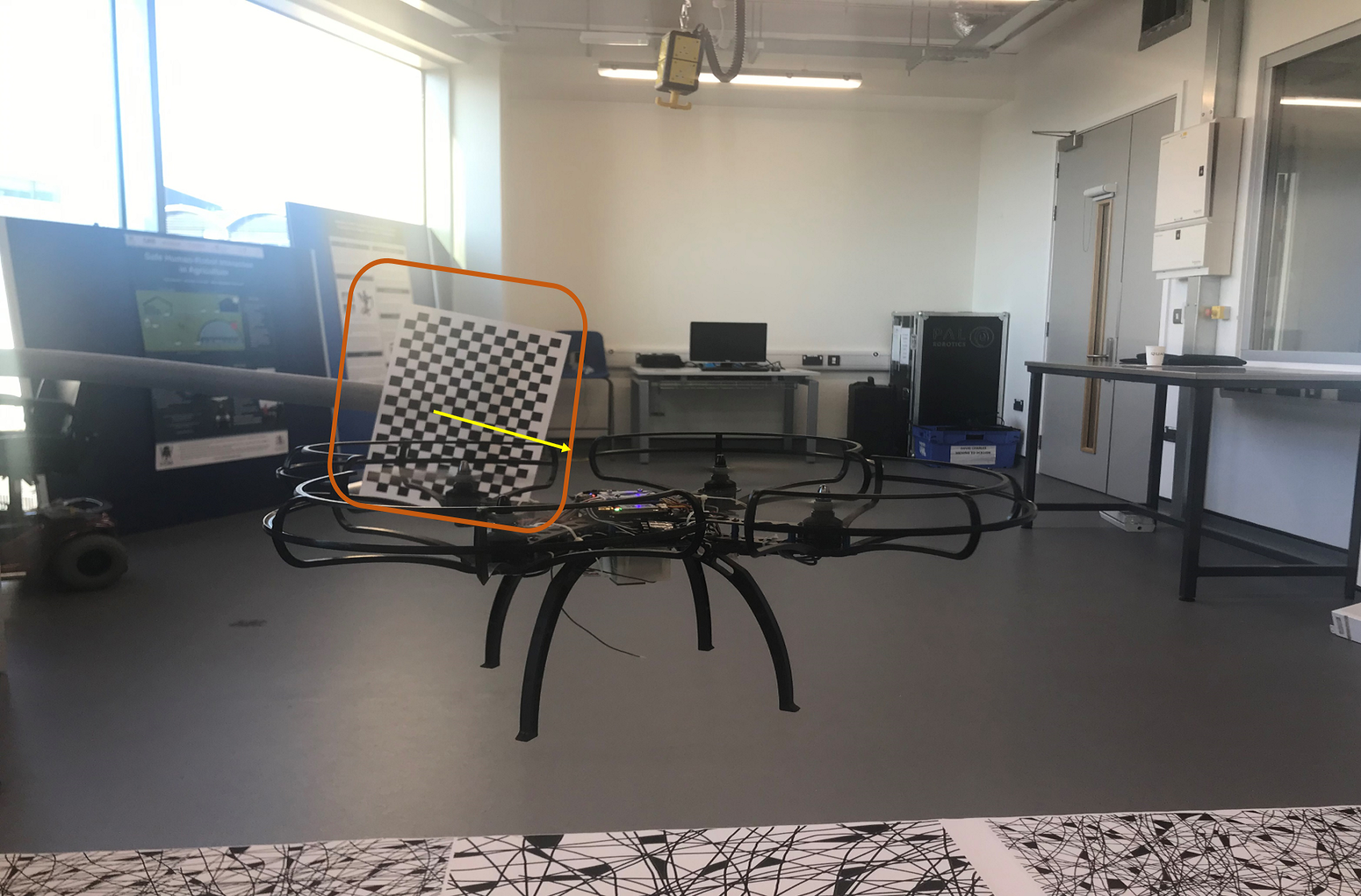}
\caption{Hovering experiments scene}\label{fig:Hovering_scene}
\end{figure}

\begin{figure}
\begin{minipage}{7cm}
\centering 
    \subfigure[Average membrane potential(Object from upside down)]
    {
      \includegraphics[width=2.5in]{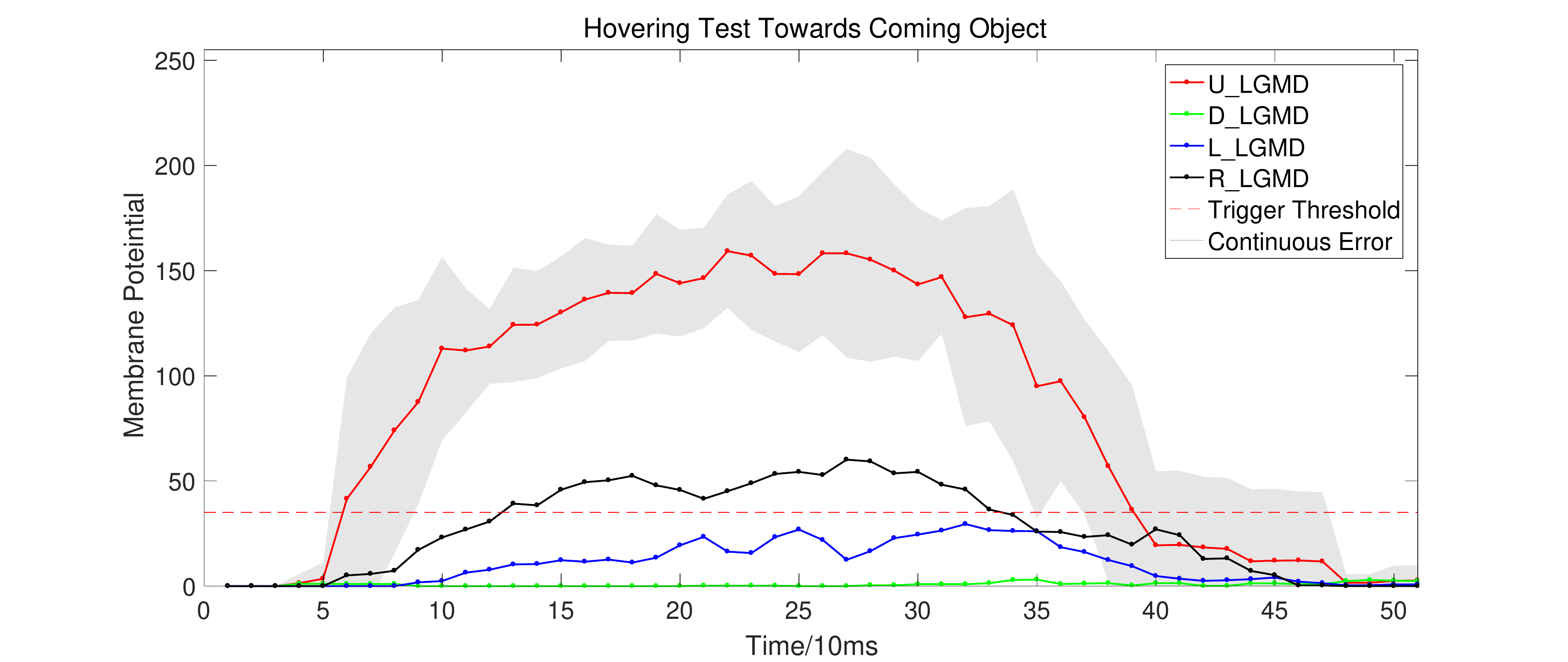}
    }
\end{minipage}
\begin{minipage}{7cm}
\centering 
    \subfigure[Average membrane potential(Object from downside up)]
    {
      \includegraphics[width=2.5in]{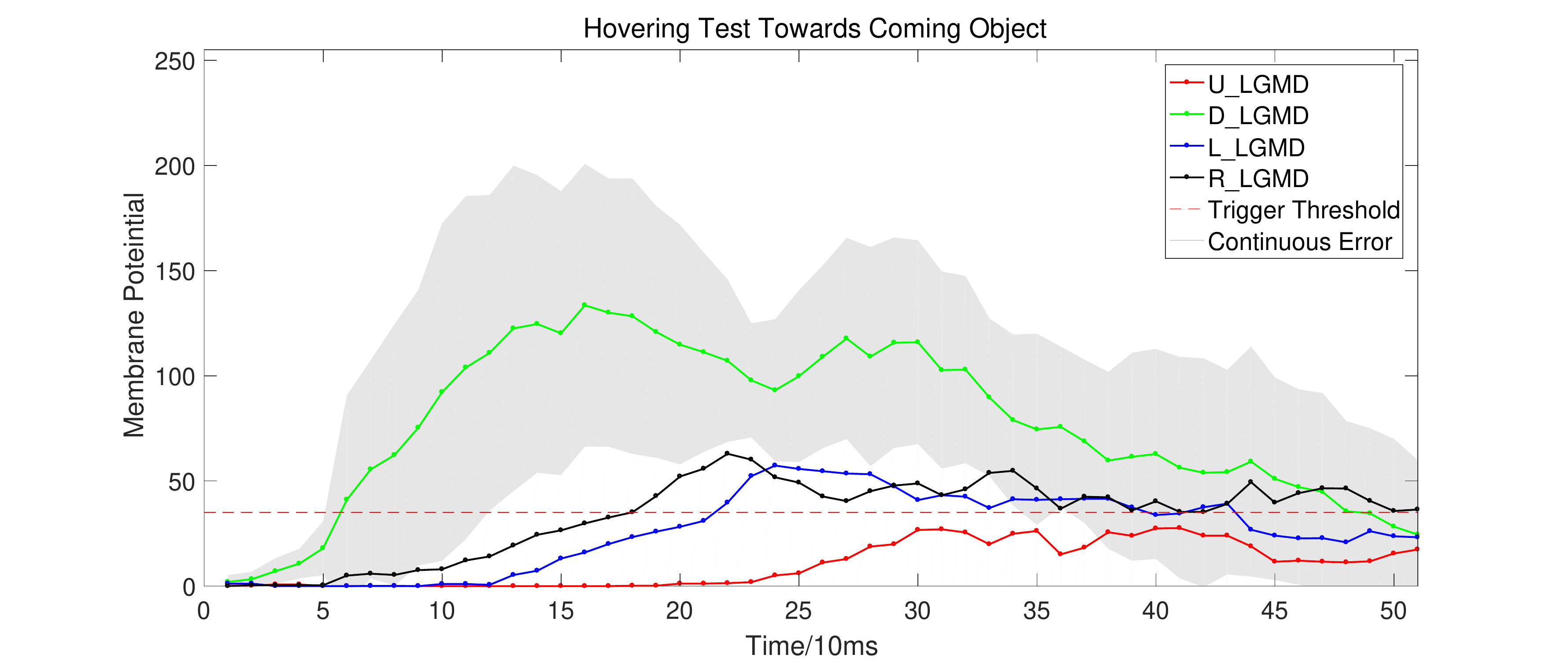}
    }
\end{minipage}
\begin{minipage}{7cm}
\centering 
    \subfigure[Average membrane potential(Object from left to right)]
    {
      \includegraphics[width=2.5in]{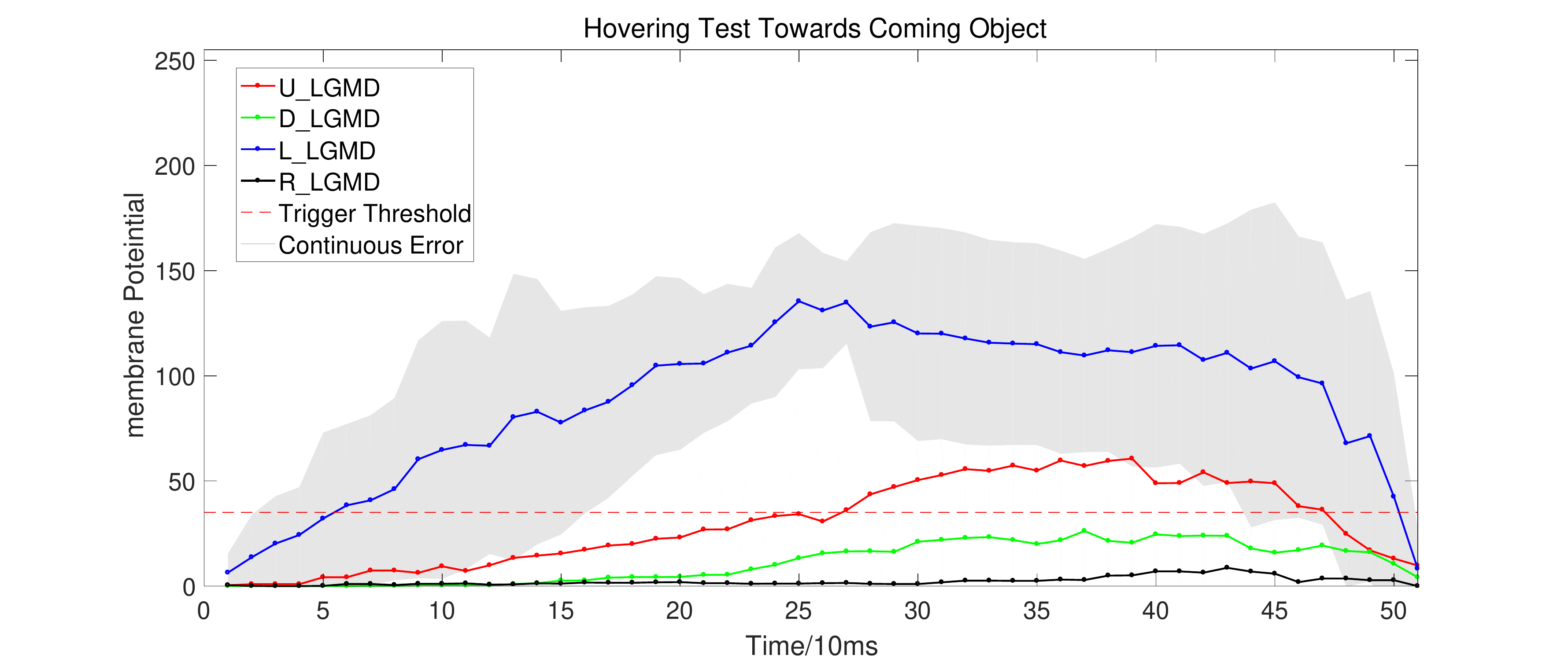}
    }
\end{minipage}
\begin{minipage}{7cm}
\centering 
    \subfigure[Average membrane potential(Object from right to left)]
    {
      \includegraphics[width=2.5in]{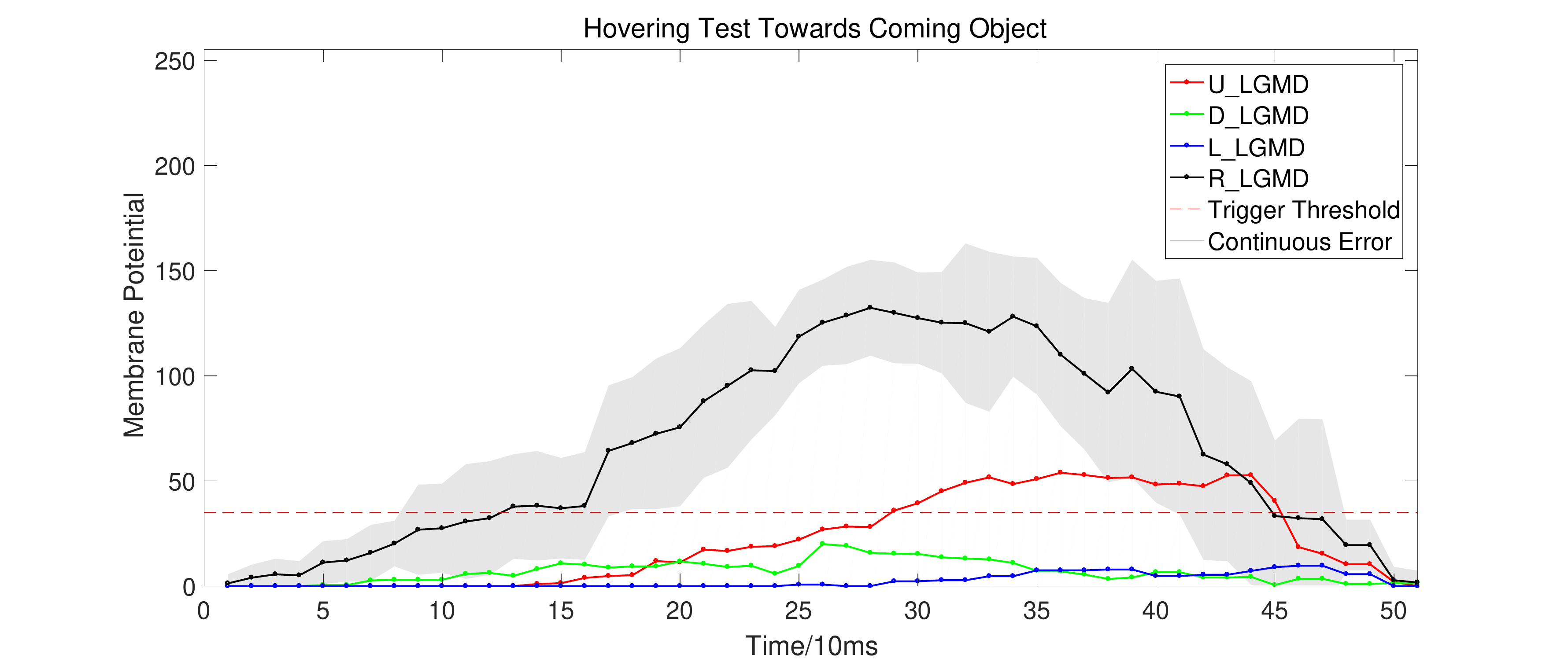}
    }
\end{minipage}
  \caption{Average membrane potential during hovering tests.(a),(b),(c),(d), reflected the average membrane potential of the four competitive LGMD neuron in trials. The shadow is the continuous error of the C-LGMD of the main direction.}\label{fig:Hovering}
\end{figure}

To further analyze the performance on quadcopter platform, we transplanted the algorithm into the embedded LGMD detector, and mounted the detector onto the quadcopter, stimulated the detector with test patterns while the quadcopter hovering in the air. Object is manually pushed towards the detector from four direction respectively, and each direction repeated 10 times.
Fig.\ref{fig:Hovering_scene} is an example of the trial scene, in which object is pushed towards the detector from left.
According to the results in Fig.\ref{fig:Hovering}, the four competitive LGMD distinguished the coming direction of the object accurately. In all the four types of trials, when LGMD exceeds its threshold, the C-LGMD indicated the main direction is leading the other average values, even if the lowest performance (lower boundary of the shadow).

\subsection{Arena Real-time Flight}
\begin{figure}
\begin{minipage}{7cm}
\centering 
  \subfigure[Left \& right object avoidance in arena test]
    {
      \includegraphics[width=2.7in]{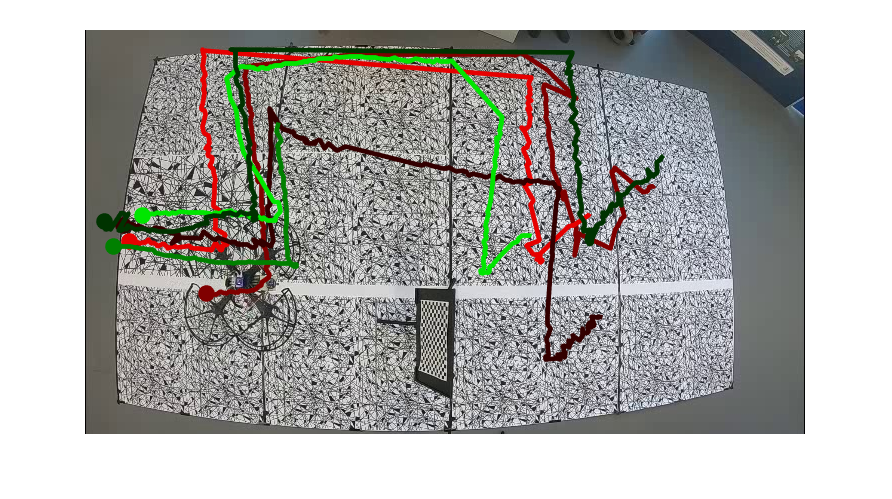}
      \label{fig:Arena_LR}
    }
\end{minipage}
\begin{minipage}{7cm}
\centering 
  \subfigure[Up \& down object avoidance in arena test]
    {
      \includegraphics[width=2.7in]{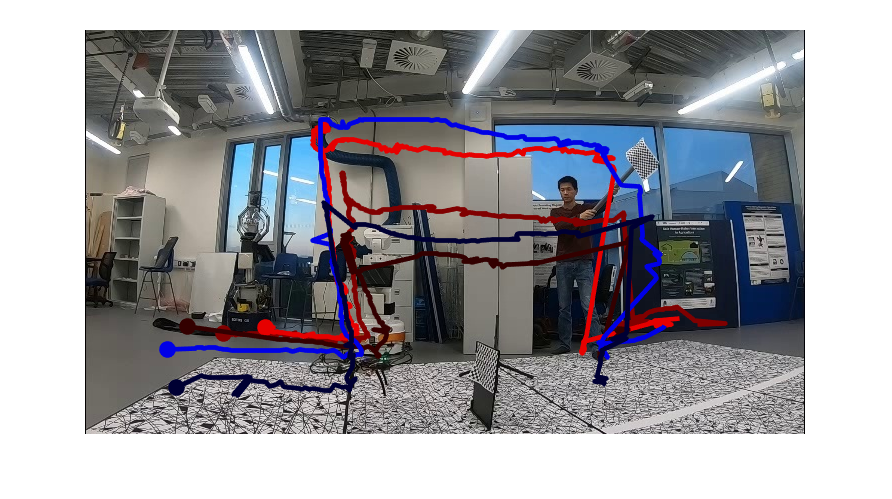}
      \label{fig:Arena_UD}
    }
\end{minipage}
  \caption{Real-time obstacle avoiding test.}\label{fig:Trajectory}
\end{figure}

Finally, real-time flight and obstacle avoidance experiments are conducted to test the performance and robustness of the proposed directionally obstacle avoiding method. Trials reflecting four directions of coming object are set in two types: obstacles on the left and right side or on the upside and downside on the UAV's route. The quadcopter is first challenged by a static obstacle and then challenged by a dynamic intruder. The results showed that the system is able to make smart escape behaviour based on the coming direction of the obstacle.
The trajectories of these trials have been extracted and overlaid on a screenshot from the video, as shown in Fig.\ref{fig:Trajectory}. Trajectories are detected by a python program using background subtractor\cite{zivkovic2004improved} and template matching\cite{lewis1995fast} method, and then printed onto a screen shot from the recorded video.

\section{Conclusion}
To conclude, a novel competitive LGMD and corresponding UAV control algorithm is proposed to address practical problems meet in UAV's LGMD application. Both simulation and realtime flight experiments were conducted to analyze the proposed method, and the results showed high robustness. Based on the proposed competitive LGMD, quadcopter's Real-time 3D collision avoidance is achieved in indoor environment.
For the future work, totally autonomous flight in a larger arena should be take to analyze the boundary of this new method.

%
%
%
%
\bibliographystyle{splncs04}
\bibliography{Conference_Cite}
%





\end{document}